%% file: egpaper_for_review.tex
\documentclass[english]{lni}

\usepackage[english]{babel}
\usepackage{color,soul}
\usepackage{csquotes}
\usepackage[
  backend=biber,
  style=LNI
]{biblatex}
\usepackage{array, makecell} %
\usepackage{tabularx,booktabs}
\usepackage{graphicx}
\usepackage{fancyhdr}
\usepackage{listings} 
\usepackage{changepage} 
\usepackage[figurename=Fig., tablename=Tab., small]{caption}
\usepackage{color}
\usepackage{xcolor}
\usepackage{footmisc}
\usepackage{url}
\usepackage{booktabs}
\usepackage{times}
\usepackage{verbatim}
\usepackage{subcaption}

\fancypagestyle{titlepage}{}

\renewcommand{\headrulewidth}{0.4pt} 
\input{math_commands.tex}

\def\etal{\emph{et~al}.}

\author{
Pedro C. Neto\footnote{INESC TEC, Porto, Portugal, pedro.d.carneiro@inesctec.pt} $^{, }$\footnote{Faculdade de Engenharia da Universidade do Porto, Porto, Portugal} ,
Fadi Boutros\footnote{Fraunhofer Institute for Computer Graphics Research IGD, Germany} $^{, }$\footnote{TU Darmstadt, Germany}, João Ribeiro Pinto$^{1, 2}$, Mohsen Saffari$^{1, 2}$,\\ Naser Damer$^{3,4}$ , Ana F. Sequeira $^{1}$, Jaime~S.~Cardoso$^{1, 2}$
}

\title{
My Eyes Are Up Here: Promoting Focus on Uncovered Regions in Masked Face Recognition
}

\begin{document}
\maketitle

\renewcommand{\refname}{References}
\setcounter{footnote}{4} 
\thispagestyle{titlepage}
\pagestyle{fancy}
\fancyhead{} 
\fancyhead[RO]{\small Promoting Focus on Uncovered Regions in Masked Face Recognition \hspace{25pt}  \hspace{0.05cm}}
\fancyhead[LE]{\hspace{0.05cm}\small  \hspace{25pt} Pedro C. Neto~\etal}
\fancyfoot{} 
\renewcommand{\headrulewidth}{0.4pt} 



\begin{abstract}
{The recent Covid-19 pandemic and the fact that wearing masks in public is now mandatory in several countries, created challenges in the use of face recognition systems (FRS). In this work, we address the challenge of masked face recognition (MFR) and focus on evaluating the verification performance in FRS when verifying masked vs unmasked faces compared to verifying only unmasked faces. We propose a methodology that combines the traditional triplet loss and the mean squared error (MSE) intending to improve the robustness of an MFR system in the masked-unmasked comparison mode. The results obtained by our proposed method show improvements in a detailed step-wise ablation study. The conducted study showed significant performance gains induced by our proposed training paradigm and modified triplet loss on two evaluation databases. }
\end{abstract}

\section{Introduction}
Computer vision tools have been successfully applied to face recognition (FR) in the past~\cite{Schroff}. New challenging conditions, such as the face occlusion caused by the use of face masks in public, mandatory during the SarsCov2 pandemic, raised limitations for well-performing and established FR methods. The pandemic has also stressed the importance of hygienic and contactless biometrics \cite{DBLP:journals/corr/abs-2102-09258}, such as FR. {Recently, the National Institute of Standards and Technology (NIST), in the scope of the ongoing Face Recognition Vendor Test (FRVT), published a study on the effect of face masks on the performance of vendor's FR systems (FRVT -Part 6A). The NIST study concluded that the algorithm accuracy with masked faces declined substantially. The Department of Homeland Security has conducted an evaluation with similar goals, however on more realistic data\footnote{https://mdtf.org/
Rally2020/Results2020}. They also observed the significant negative effect of wearing masks on the accuracy of automatic FR methods.} 

The lack of robustness of current systems to perform masked face recognition (MFR) fostered an interest in the research community to address this challenge~\cite{li2021cropping,hong2020masked,geng2020masked,ding2020masked}. Damer \etal~\cite{DamerBiosig2020} evaluated the verification performance drop in three face biometric systems when verifying masked vs not-masked faces compared to verifying not-masked faces to each other. This study was extended~\cite{DamerIetbmt2021} to both synthetic and real masks, pointing out the questionable use of simulated masks to represent the real mask effect on face recognition. Furthermore, the performance of human inspectors in face verification has been shown to have a drop, consistent with automatic FR systems, when faces are masked \cite{DBLP:journals/corr/abs-2103-01924}. The effect of facial masks extends to other components of biometric systems as it has been shown to largely change the behaviour of face presentation attack detection \cite{DBLP:journals/corr/abs-2103-01546}.  Recently, Boutros \etal~\cite{boutros2021unmasking} proposed a template unmasking approach that can be adapted on top of any face recognition network aiming at creating unmasked-like templates from masked faces by the proposed self-restrained triplet loss. {Other initiatives were also incited by the lack of systems capable of handling this task, such as the ``Competition on Masked Face Recognition'' (IJCB-MFR-2021 \cite{DBLP:conf/icb/BoutrosDKRKRKFZ21}) and ``The International Workshop on Face and Gesture Analysis for COVID-19'' (FG4COVID19)\footnote{\url{https://fg4covid19.github.io/index.html}}.} 

{The work proposed in this paper comprises the construction of a synthetic masked face dataset based on the VGGFace2~\cite{Cao2018} and proposes a solution to address the challenge of MFR. This solution is based on a proposed loss that combines the traditional triplet loss and the mean squared error (MSE) intending to improve robustness in the comparison between masked and unmasked samples}.  

The contributions of this work are: 
1) A cascaded training paradigm that leverages the benefits of both a conventional identity classification learning in the first stage and the subsequent embedding optimization fine-tuning stage.
2) A specifically modified triplet loss function (for the embedding optimization) that incorporates a mean square error measurement to control the training process in a weighted manner.
3) A thorough ablation study on multiple databases (including our own created database), showing, in a step-wise manner, the benefit of our training paradigm and the specifically designed loss.

This paper is organised as follows. In this introductory section, we contextualise the challenge addressed within the related work and detail the contributions of the paper; and in the conclusion, we reflect on the findings and future work possibilities. In Section~\ref{methods} we present the methodology used. Section~\ref{esr} details the metrics, the datasets used (the creation of the synthetic face masks and the dataset with real masks), the implementation details, and finally, presents the results and its discussion.

\section{Methodology}
\label{methods}

{Performing the direct optimization of embedding predictions is often trickier than learning a model capable of performing classification. Hence, we propose a constrained triplet loss, specially crafted for masked face recognition (MFR), to be used after the classification optimization. Our approach redirects the focus of learned embeddings towards unmasked areas. In the embedding learning stage, the training focuses on comparing two images against a reference (anchor) and distinguishing between images of the same person (positives) and from a different person (negatives). Together, the anchor, the positive, and the negative form a triplet. This will enable the model to capture the benefits of both strategies and therefore be more flexible to inter-class variations, as we will experimentally illustrate.

In this section, we start by describing the approach followed to train these models for classification. Afterwards, we expose our proposed change to the triplet loss that improves the representations learned by these models for MFR.}

\paragraph{Classification Training:}
In our scenario, the number of classes and the universe of possible inputs is unknown once the model is deployed. Thus, supervised classification methods can only guide us up to a certain point. Nonetheless, they have been used as a technique to improve the convergence speed of the model for other tasks. Therefore, initially, the problem is approached as a closed-set recognition. Afterwards, the pre-trained model is fine-tuned towards learning meaningful embeddings. 

The approach to train the classification model was based on minimizing the cross-entropy (CE). This loss function is frequently used for the classification of the input as a single class, which suits our use case since a picture can only belong to one subject. It attempts to minimize the confidence of the model on erroneous classes while maximizing its confidence in the correct class. The validation occurred after each epoch and it evaluated the accuracy of the model in the classification of masked pictures unseen during the training. And while these images were unseen, the network already knew the subject from past pictures. To separate validation and training sets, we followed an $80\% / 20\%$ data split. This training process is similar to the one designed by Cao~\etal~\cite{Cao2018}.

\paragraph{Embedding Optimization:} Embedding optimization is a task that requires the network to learn the representations of the inputs instead of classifying them. This is no trivial task, and the hyperparameters search of this process has to be done carefully since this is an expensive task when compared to the use of classification losses~\cite{Yuan2020}. For this, the fully connected layer is removed, and another untrained embedding layer is added. Besides this last layer, all the weights of the network are frozen, and thus, further training does not update them. To train the embedding layer, which outputs an embedding vector with a size of 512, the triplet loss is used, based on Equation~\ref{eq:tl}.  
\begin{equation} \label{eq:tl}
Triplet Loss=\sum_{a,p,n}{max(0,\alpha - || x_a - x_n||_2^2 + || x_a - x_p||_2^2 )}
\end{equation}
\vspace{-0.2cm}
\begin{equation} \label{eq:4}
x_i = W' \frac{\phi(l_i))}{||\phi(l_i)||_2} 
\end{equation}

Equation~\ref{eq:4} describes the embedding layer added after the last convolutional layer. It receives as input the output of the convolutional layer (represented by $l$ on the equation) and normalizes it in the euclidean space. 

The triplet loss has some aspects that serve our goals, for instance, it relies on three inputs, referred to as the ``anchor'', the ``positive'' and the ``negative'', which suits the structure of our evaluation method. Moreover, this loss verifies the distances between the anchor and the positive and between the anchor and the negative. It penalizes the network if the last one is smaller. The formulation of the triplet loss is given by Equation~\ref{eq:tl}, where it is possible to see that it penalizes the model if the distance between the negative and the anchor is shorter than the one of the positive and the anchor. Moreover, the equation includes a term $\alpha$, which is the margin. The margin, which in this case is set to 0.2, helps the model to define some separability between positives and negatives.


\begin{figure}[h!]
   \centering
    \includegraphics[width=0.75\linewidth]{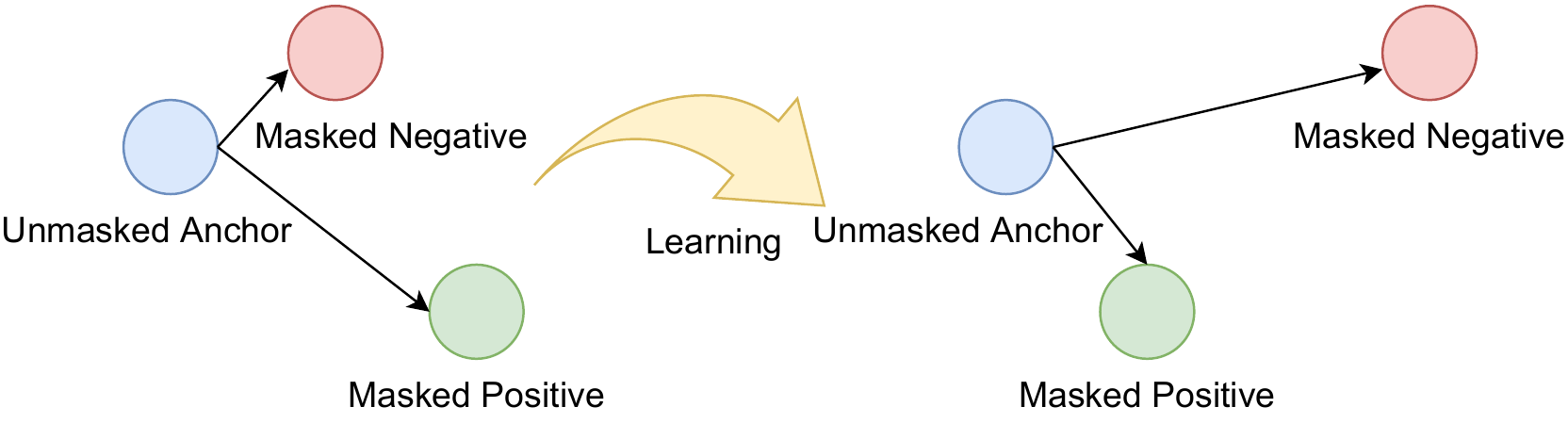}
  \caption{Triplet loss effect in the euclidean space}
  \label{triplet_objective} 
\end{figure}

\begin{figure}[h!]
    \centering
   \includegraphics[width=0.8\linewidth]{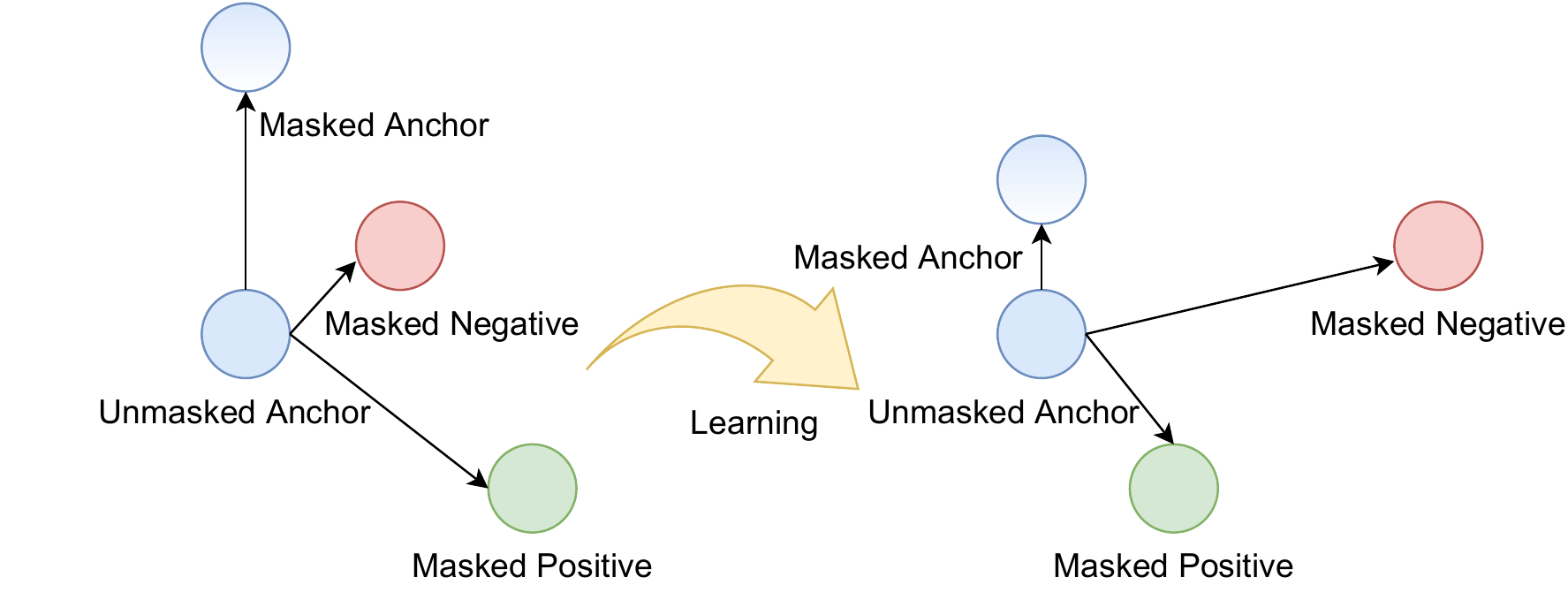}
  \caption{Proposed Triplet loss effect in the euclidean space}
  \label{triplet_objective_new} 
\end{figure}

It is possible to see on Figure~\ref{triplet_objective} 
the effect that optimizing with triplet loss has on the embeddings. The anchor is a randomly selected image (without mask) to be used as a comparison point. The positive image, is a masked image from the same identity as the anchor image, whereas the negative is from a different identity. Our proposed approach, $TripletLoss_{Prop}$, constraints the loss of the original triplets, through the minimization of the distance between the masked and unmasked anchor embeddings. Our loss is formulated in Equation~\ref{eq:5} and the effects of its optimization are visible on 
Figure~\ref{triplet_objective_new}.

\begin{equation} \label{eq:5}
TripletLoss_{Prop}=\sum_{a,am,p,n}{Triplet Loss(a,p,n) + MSE(am,a)}
\end{equation}
Since the mask occlusions are always on the same facial area, it is known that the network should focus its attention on areas that will not have occlusions. And thus, as seen on Equation~\ref{eq:5}, a mean squared error term is added to the loss. This way, we introduce more information to the model so that embedding optimization can be done more effectively.

\section{Experimental Setup and Results}
\label{esr}
\paragraph{Evaluation Metrics:} {To report the results we present the \textit{false non-match rate (FNMR)}; the \textit{FMR100} and \textit{FMR10} which are the lowest \textit{FNMR} for a \textit{false match rate (FMR)} $< 1.0\%$ and $< 10.0\%$, respectively; the \textit{equal error rate (EER)}; and the \textit{area under the receiver operating characteristic curve (AUC)}. We also report the genuine mean (GMean) and impostors mean (IMean), which represent the mean distances between the embeddings of the same individual and from different people respectively. }


{\bf Face Data}:
{The development of face recognition methods requires large and diverse datasets. When the Covid-19 pandemic started and it became evident that the use of face masks had a negative impact on FR systems there was no ready-to-use data for research. The creation of synthetic data allows leveraging from existing data so that it fits the problem. Still, using real data is crucial as the ultimate test to the models and the community started to also collect face samples of individuals using face masks. The creation of these two types of data used in our method is described as follows.}

\begin{figure}[h!]
    \centering
  \begin{subfigure}[b]{0.15\linewidth}
       \includegraphics[width = \linewidth]{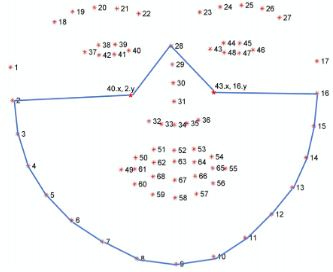}
       \caption{Wide,\\high}
  \end{subfigure}
  \begin{subfigure}[b]{0.15\linewidth}
       \includegraphics[width = \linewidth]{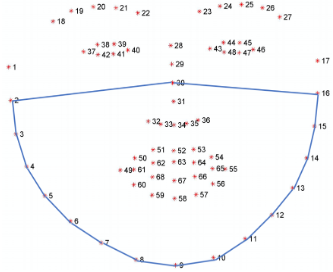}
       \caption{Wide,\\medium }
  \end{subfigure}
  \begin{subfigure}[b]{0.15\linewidth}
       \includegraphics[width = \linewidth]{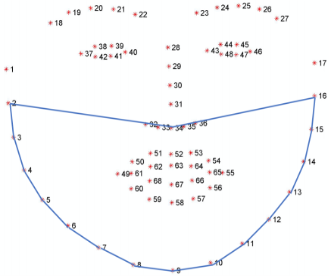}
       \caption{Wide,\\low}
  \end{subfigure}
  \begin{subfigure}[b]{0.15\linewidth}
       \includegraphics[width = \linewidth]{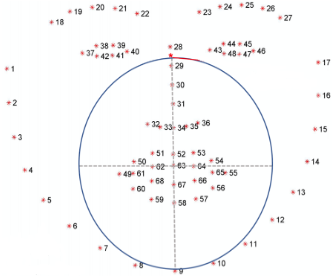}
       \caption{Round, \\high}
  \end{subfigure}
  \begin{subfigure}[b]{0.15\linewidth}
       \includegraphics[width =\linewidth]{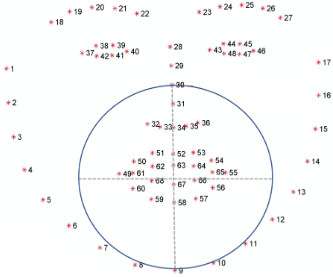}
       \caption{Round, \\medium} 
  \end{subfigure}
  \begin{subfigure}[b]{0.15\linewidth}
       \includegraphics[width =\linewidth]{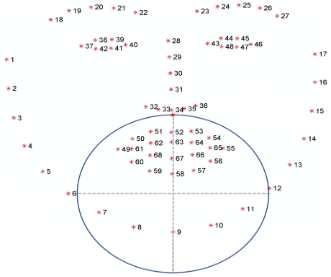}
       \caption{Round, \\low}
  \end{subfigure}
  \caption{{Examples of face landmarks obtained from one image with different types of masks added. These masks vary in shape and face coverage.}}
    \label{fig1} 

\end{figure}

\textit{Synthetic masked face data (SMFD)}:
{Here we describe the synthetic masked face data (SMFD) creation process. Adding a facial mask requires information regarding facial landmarks. Moreover, the position and inclination of the face affect the positioning of the mask. Due to the lack of large-scale pairs of masked and unmasked identities, in this work, we synthetically generate different types of masks and adjust them on the unmasked samples of the VGGFace2 dataset~\cite{Cao2018}. The dataset includes 3,310,000 face images from 8,631 train identities and 500 disjoint identities for the test, and includes a diverse set of samples regarding the various poses, ages, and ethnicity. 
Mask generation is carried out using the proposed algorithm by NIST~\cite{ngan2020ongoing}. The algorithm exploits the Dlib C++ toolkit to obtain 68 facial landmarks for each image; afterward, using the extracted facial landmarks and interpolation between the points, various synthetic masks are generated. The details of the landmark extraction and mask generation are described in~\cite{ngan2020ongoing,king2009dlib}. Due to variability regarding shape (Wide vs. Round) and face coverage (High, Medium, Low) we obtained six possible combinations. Figure \ref{fig1} shows the result of applying the mask generation algorithm on a randomly selected sample from VGGFace2.  } Both the shape of the mask and its colour are randomly selected, for each image, while generating the masks.

\textit{Real masked face dataset}:
We evaluated our proposed solution using masked face recognition competition dataset (MFRC-21)~\cite{DBLP:conf/icb/BoutrosDKRKRKFZ21}. MFRC-21 dataset was collected on 3 different days from 47 subjects. The first day is considered as a reference session, while the second and third sessions are considered probe sessions. 
In each session, 3  (two with mask and one without) videos are recorded using a webcam, while the subjects are requested to look directly into their camera. An overlapping database, and the same capture and frame selection procedure is describe  in~\cite{DamerBiosig2020,DamerIetbmt2021}. In total, the references contain 470 unmasked images and 940 masked images. The probes contain 940 unmasked images and 1880 masked images. We evaluate our proposed solution under two evaluation scenarios.
The first is between unmasked references and masked probes (U-M) and the second is between masked references and masked probes (M-M).


\begin{figure}[h!]
    \centering
  \begin{subfigure}[b]{0.49\linewidth}
       \includegraphics[width = \linewidth]{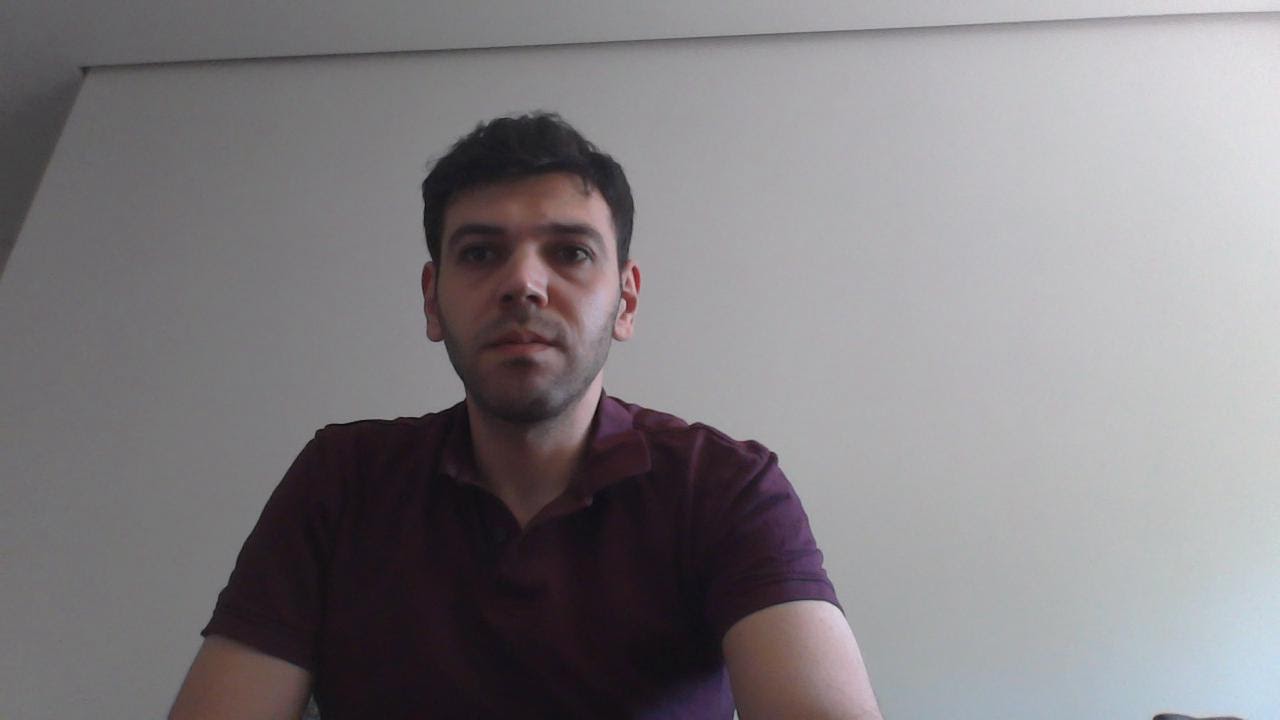}
       \caption{Example from the test set of a reference image without a mask}
       \label{unmaskedfadi}
  \end{subfigure}
  \begin{subfigure}[b]{0.49\linewidth}
       \includegraphics[width = \linewidth]{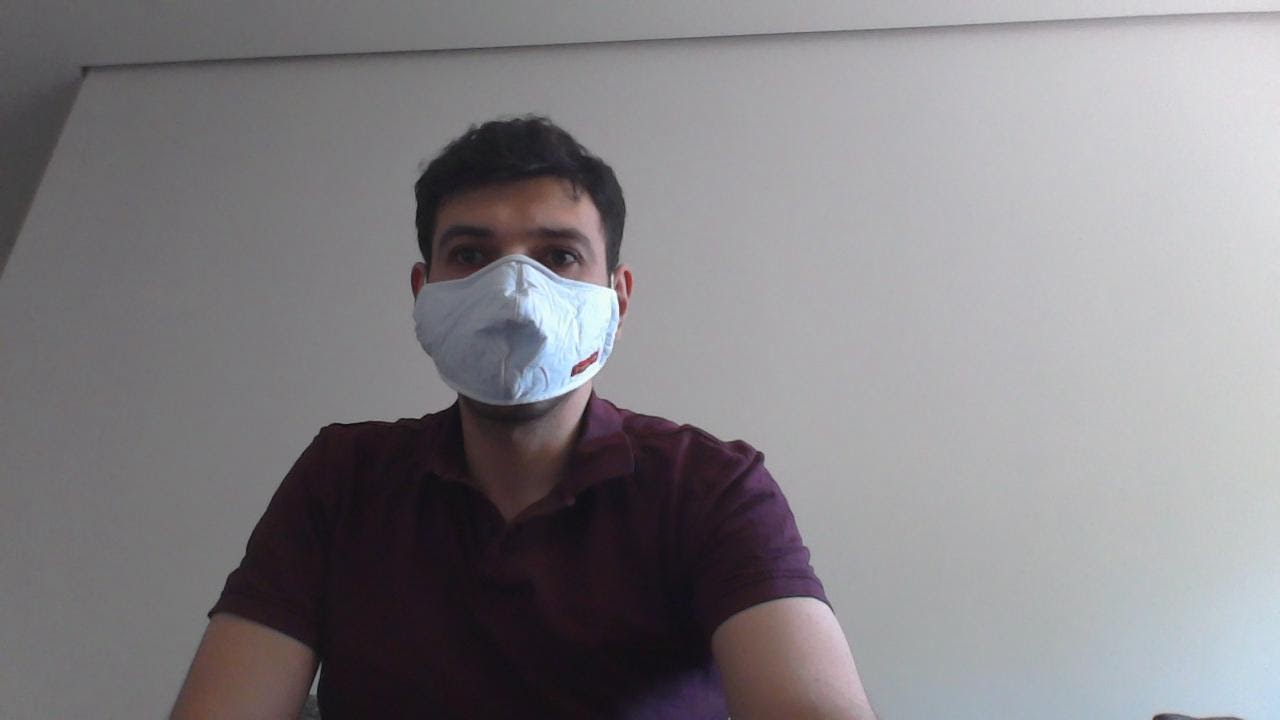}
       \caption{Example from the test set of a probe image with a mask}
       \label{maskedfadi}
  \end{subfigure}

  \label{fig2} 
  \caption{{Examples of images from the real masked faces dataset: images of the same individual with (a) and without a mask  (b) (from~\cite{DamerBiosig2020}).}}
\end{figure}

{\bf Implementation Details}: {To implement the mask FR model, we exploit ResNet50 architecture~\cite{he2016deep} to extract the features from masked/unmasked facial samples. We trained the model by making use of cross-entropy (CE). The training of this model required around 150 thousand iterations. It trained with an initial learning rate of 0.1. It was decreased by a factor of 10 whenever the validation accuracy decreased. Stochastic gradient descent was used, with a batch size of 400, momentum of 0.9, and 0.0005 weight decay. We did not use any face alignment of the VGGFace2 dataset, giving as input of the model 224x224 images. After achieving convergence, it was fine-tuned with triplet loss and the combination of triplet loss with the mean squared error. Triplets were randomly generated at training time, and thus, it was unlikely to have them seen by the network more than once. Furthermore, we did not use any triplet mining. The models trained for 65 thousand iterations with a batch size of 200. And thus, 13 million triplets were created and used for the weight updates. The margin hyper-parameter $\alpha$ in Triplet loss is empirically determined as $0.2$.}


{\bf {Results and discussion}}: {We evaluated our method on two distinct datasets. One evaluation used synthetic masked face data (SMFD), with all the identities used for testing being disjoint from the training identities. We also evaluated the model with real masked face data (RMFD). Evaluating on these datasets allows us to infer the generalization capabilities
of the model for unknown identities and images with different characteristics from the gallery images (e.g.real masks). The results are provided using the already mentioned metrics: GMean, IMean, AUC, EER, FMR100, and FMR10.}

{Our method is evaluated through a detailed step-wise ablation study that allowed us to understand the impact of the proposed modification of the triplet loss. Hence, we evaluate the model, in both datasets, with different training frameworks. This allows us to capture information regarding the impact of individual components of the model, such as, training with triplet loss, or not optimizing the embeddings. Besides, we also included the results of the method referred to in the tables as ``VGG Face''~\cite{Cao2018, Schroff} consisting of an Inception-ResNet pre-trained on the original VGGFace2 datasets.}

{In Table~\ref{res-smfd}, it can be observed that good performance is achieved with just cross-entropy training (CE). Nevertheless, optimizing the produced embeddings with triplet loss (CE+TL) led to significant improvements in performance, lower distances for impostors, and higher distances for genuines. Moreover, our proposed adapted triplet loss resulting from the addition of the MSE constraint (CE+TL+MSE) lead to even more significant improvements, for example, approaching the AUC to $0.99$, besides improvements in all the other metrics. 

\begin{table*}
 \caption{Results obtained for the synthetic masked face data (EER, FMR100, FMR10 in $\%$).}
\label{res-smfd}
\centering
\begin{tabular}{lcccccc}
\toprule
Method      & GMean & IMean & AUC & EER & FMR100 & FMR10 \\ 
\midrule
 VGG Face\cite{Cao2018,Schroff}  & 0.505 & 0.325 & 0.951 & 11.8 & 38.2 & 13.5 \\
 CE Loss & 0.528 & 0.426 & 0.941 & 13.2 & 38.5 & 21.5\\
 CE + TL & 0.601 & 0.320 & 0.977 & 7.8 & 28.9 & 11.9\\
 CE + TL + MSE \textbf{(Ours)}   & 0.596 & 0.319 & \textbf{0.985} & \textbf{6.2} & \textbf{18.5} & \textbf{4.1} \\
\bottomrule
\end{tabular}
\end{table*}

In Table~\ref{res-rmfd}, can be observed that, the model's performance degrades when compared to the previous table. Regardless of that, for the targeted comparison mode - U versus M - the results show the superior performance of our proposed loss. The distance of impostors increases as the distance of genuines increases too. Furthermore, it is possible to conclude that the model is competent in the task despite being trained only on synthetic data.}

\begin{table*}
 \caption{Results obtained for the real masked face data (in the column ``Mode'': U and M stands for unmasked and masked data, respectively; EER, FMR100, FMR10 in $\%$).}
\label{res-rmfd}
\centering
\begin{tabular}{lccccccc}
\toprule
Method      & Mode & GMean & IMean & AUC & EER & FMR100 & FMR10 \\ 
\midrule
 VGG Face\cite{Cao2018,Schroff} & \makecell{U-M\\M-M}& \makecell{0.523\\0.616} & \makecell{0.426\\0.461}  &  \makecell{0.769\\0.847}  & \makecell{29.419\\23.552}  & \makecell{90.587\\68.979} & \makecell{58.959\\38.159}   \\
 \cline{1-8}
 CE Loss & \makecell{U-M\\M-M} & \makecell{0.610\\0.702} & \makecell{0.475\\0.503} & \makecell{0.931\\0.936}& \makecell{11.687\\\textbf{9.002}} & \makecell{32.041\\\textbf{16.628}} & \makecell{12.852\\\textbf{8.791}}\\
 \cline{1-8}
 CE + TL& \makecell{U-M\\M-M} & \makecell{0.647\\0.699} & \makecell{0.396\\0.414}& \makecell{0.943\\0.945}& \makecell{11.213\\10.806}  & \makecell{34.744\\26.457} & \makecell{11.874\\11.249}\\
 \cline{1-8}
 CE + TL + MSE \textbf{(Ours)} & \makecell{U-M\\M-M}  & \makecell{0.649\\0.699} & \makecell{0.383\\0.390}& \makecell{\textbf{0.957}\\\textbf{0.959}}& \makecell{\textbf{9.799}\\9.292} & \makecell{\textbf{28.252}\\23,507} & \makecell{\textbf{9.678}\\9.035}  \\
\bottomrule
\end{tabular}
\end{table*}

{It  should be noted that, the M-M mode experimental results do not keep up with the U-M one, in order words while our method improves the U-M, it offers no improvements for M-M recognition. This can be due to the fact that the embedding optimization process is made in a way that the model is trained to minimise the distance between masked and unmasked (U-M) genuine pairs, thus aiming at making it greater than the distance between imposter pairs. However, this was performed because the main application scenario commonly would contain unmasked references, such as automatic border control with an unmasked passport-stored reference and a possible masked live probe.}

\begin{figure}[h!]
    \centering
  \begin{subfigure}[b]{0.13\linewidth}
       \includegraphics[width = \linewidth,height=2.4cm]{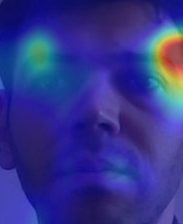}
       \caption{}
  \end{subfigure}
  \begin{subfigure}[b]{0.13\linewidth}
       \includegraphics[width = \linewidth,height=2.4cm]{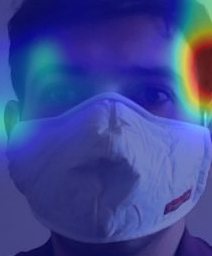}
       \caption{}
  \end{subfigure}
  \begin{subfigure}[b]{0.13\linewidth}
       \includegraphics[width = \linewidth,height=2.4cm]{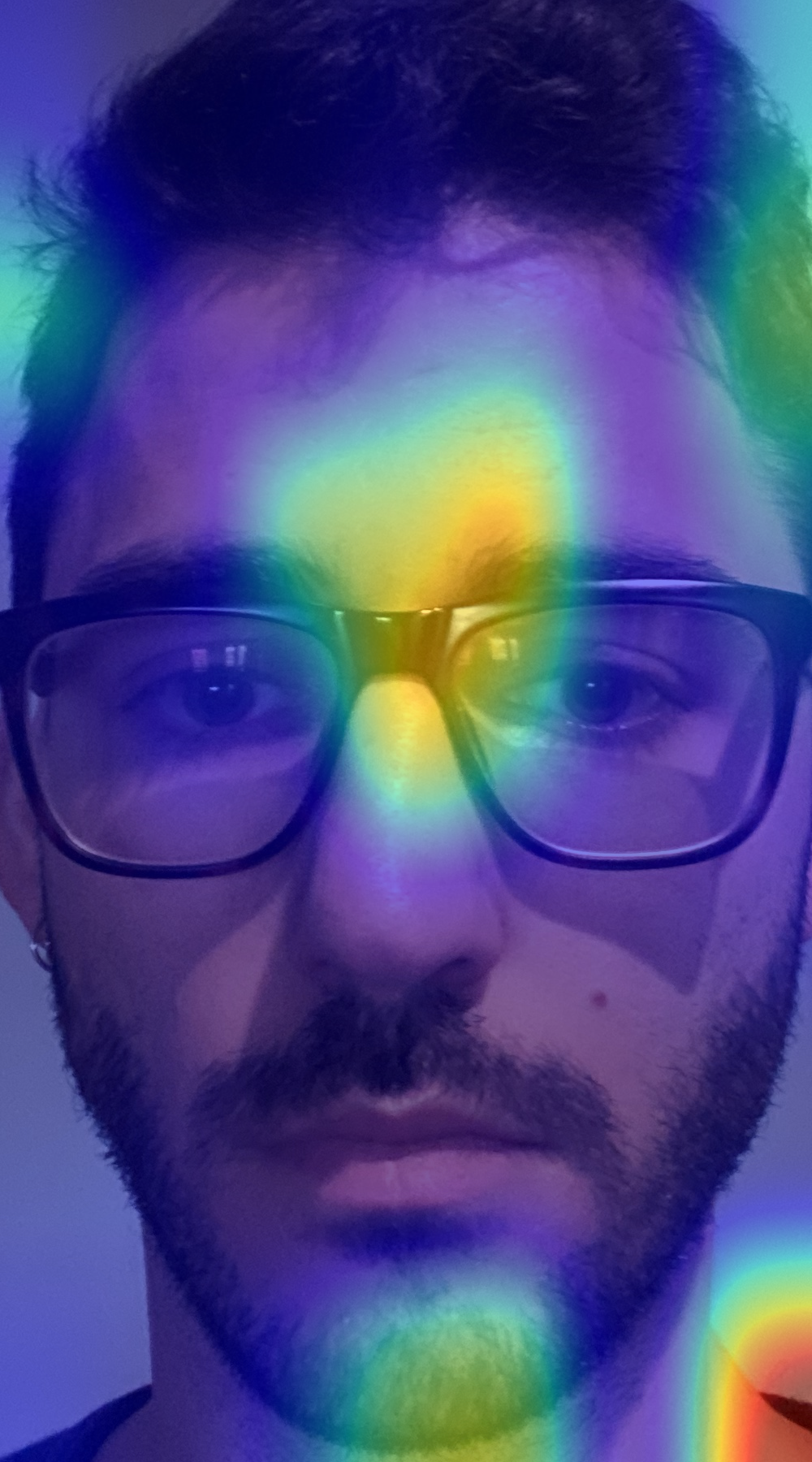}
       \caption{}
  \end{subfigure}
  \begin{subfigure}[b]{0.13\linewidth}
       \includegraphics[width = \linewidth,height=2.4cm]{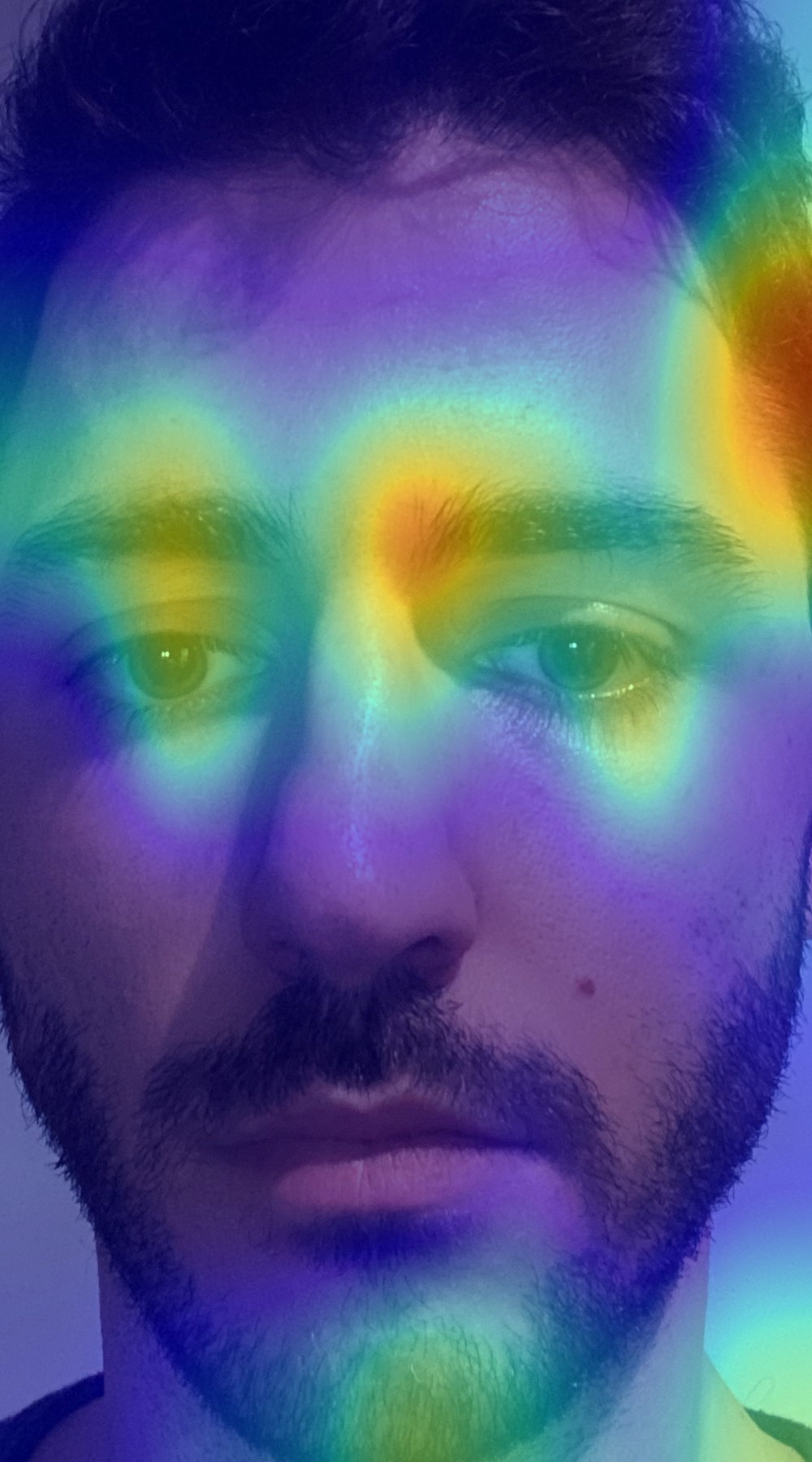}
       \caption{}
  \end{subfigure}
  \begin{subfigure}[b]{0.13\linewidth}
       \includegraphics[width = \linewidth,height=2.4cm]{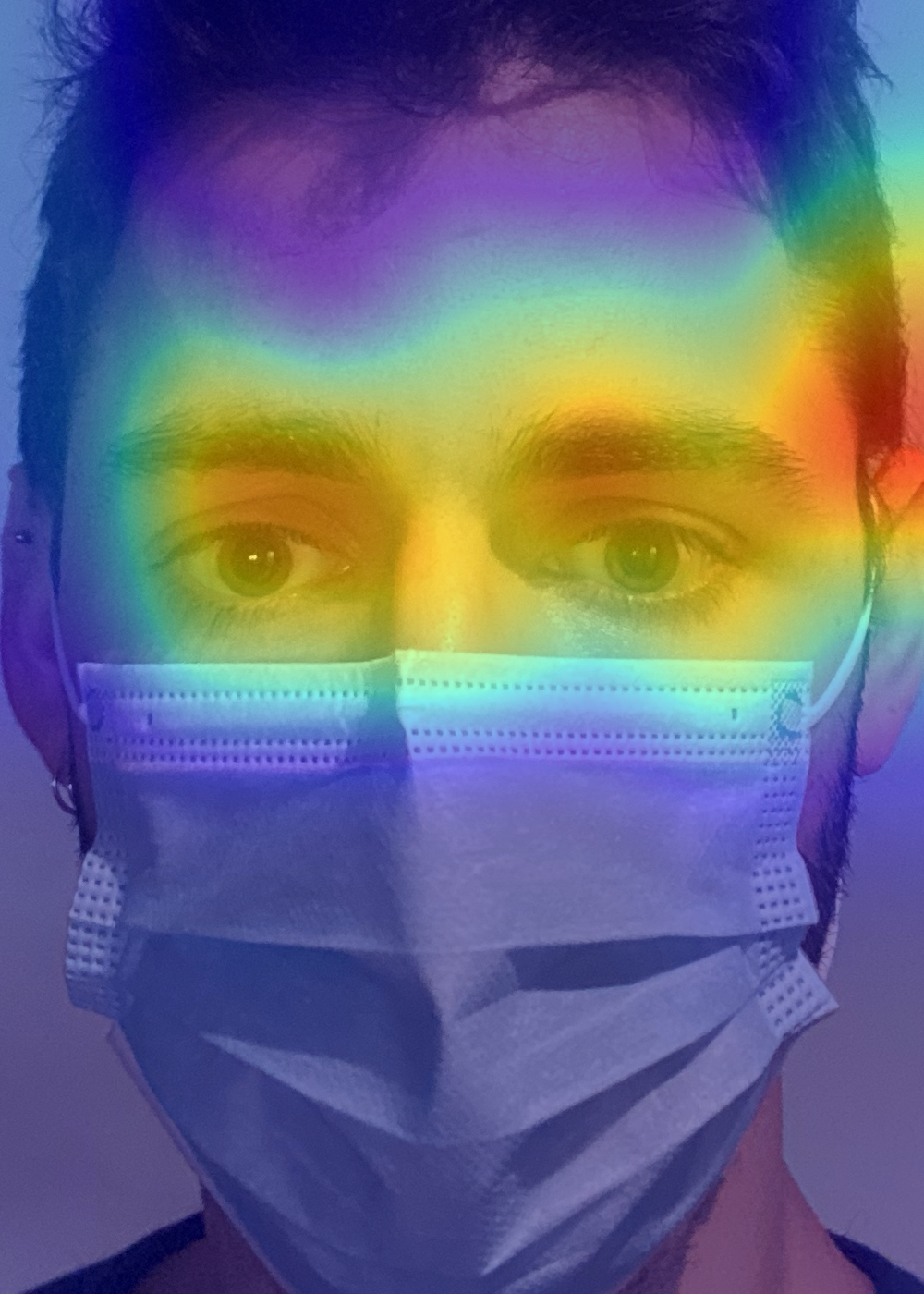}
       \caption{}
  \end{subfigure}
  \begin{subfigure}[b]{0.13\linewidth}
       \includegraphics[width = \linewidth,height=2.4cm]{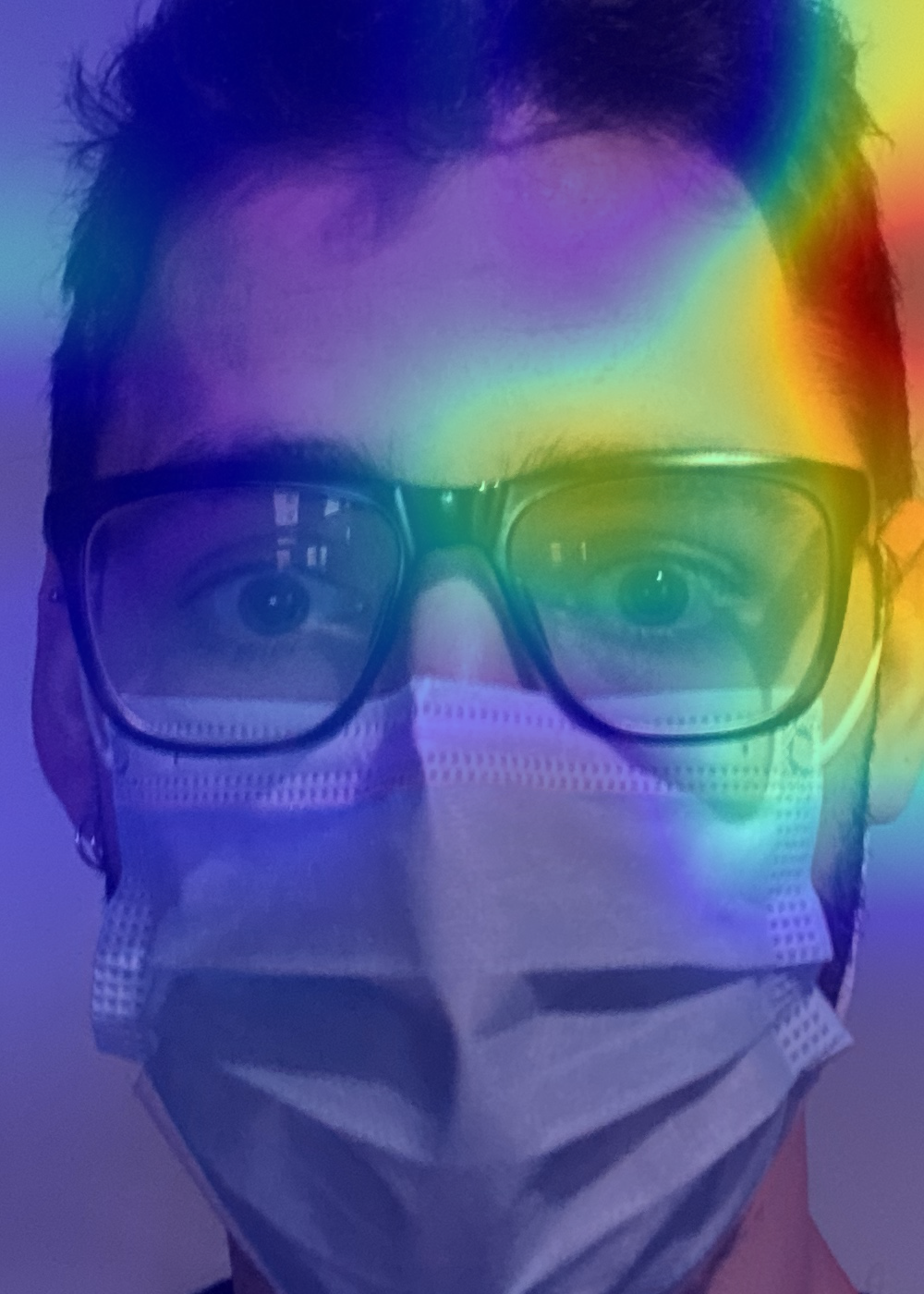}
       \caption{}
  \end{subfigure}
  
  \begin{subfigure}[b]{0.13\linewidth}
       \includegraphics[width = \linewidth,height=2.4cm]{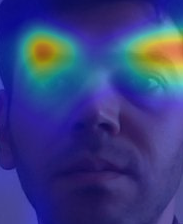}
       \caption{}
  \end{subfigure}
  \begin{subfigure}[b]{0.13\linewidth}
       \includegraphics[width = \linewidth,height=2.4cm]{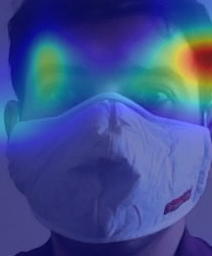}
       \caption{}
  \end{subfigure}
  \begin{subfigure}[b]{0.13\linewidth}
       \includegraphics[width = \linewidth,height=2.4cm]{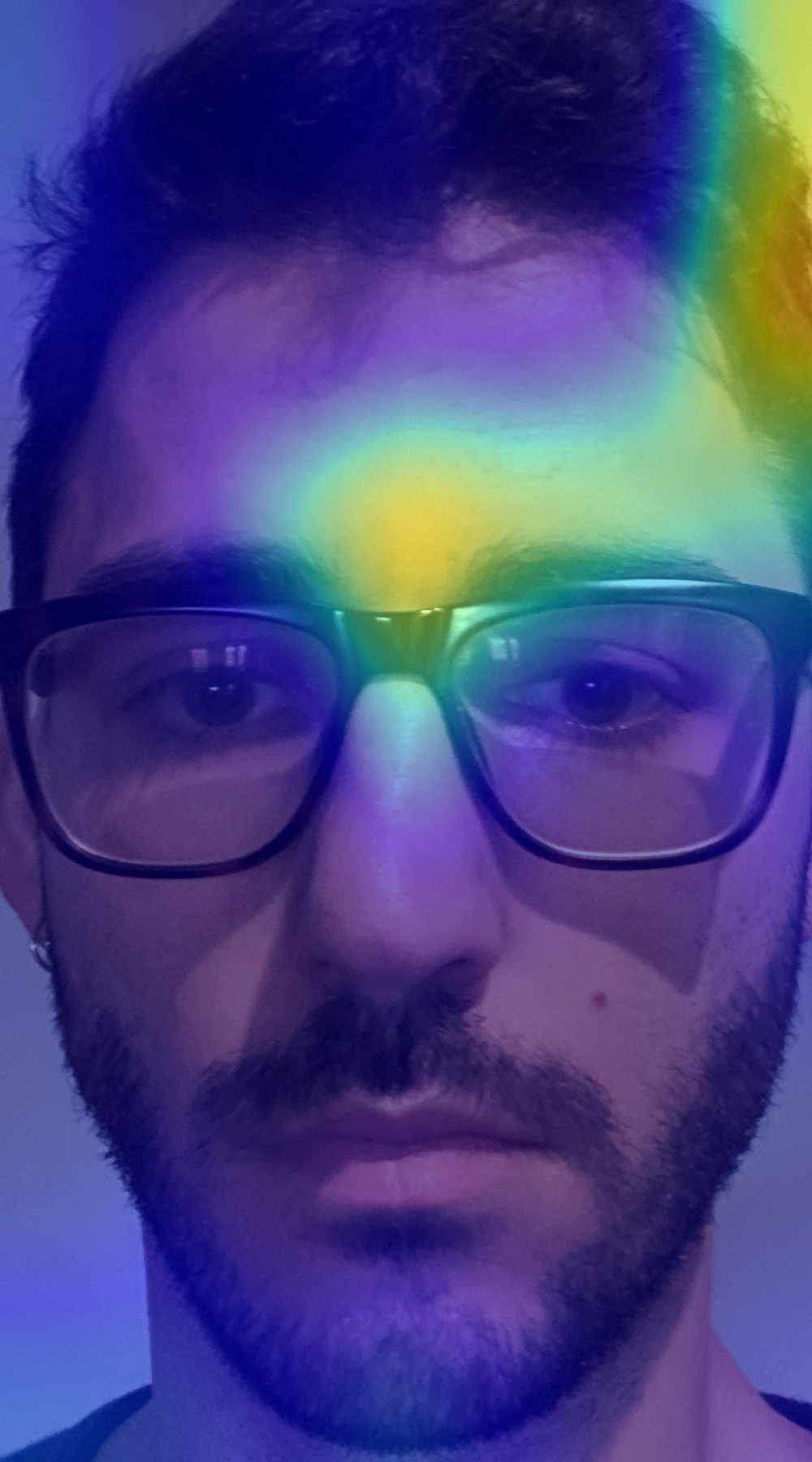}
       \caption{}
  \end{subfigure}
  \begin{subfigure}[b]{0.13\linewidth}
       \includegraphics[width = \linewidth,height=2.4cm]{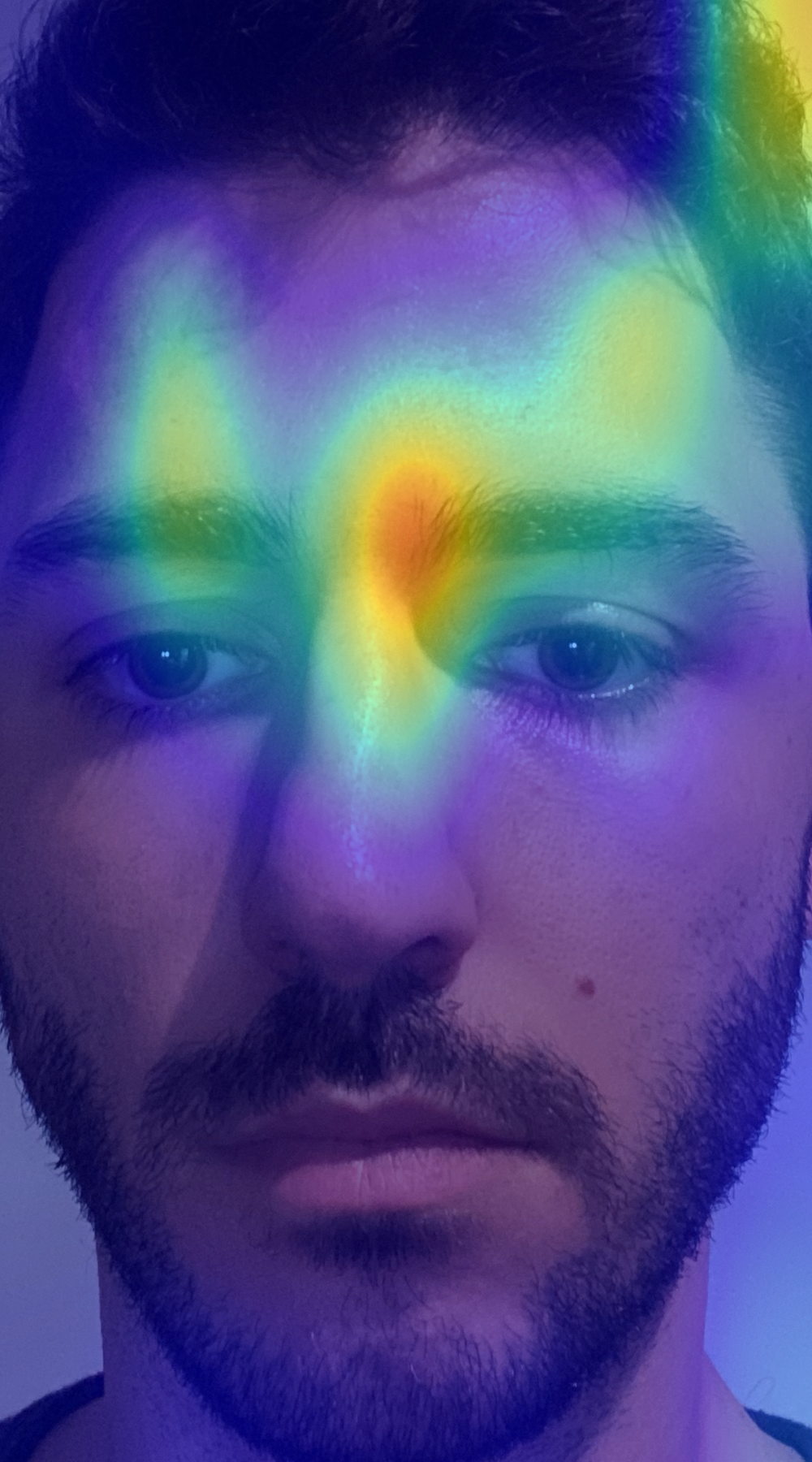}
       \caption{}
  \end{subfigure}
  \begin{subfigure}[b]{0.13\linewidth}
       \includegraphics[width = \linewidth,height=2.4cm]{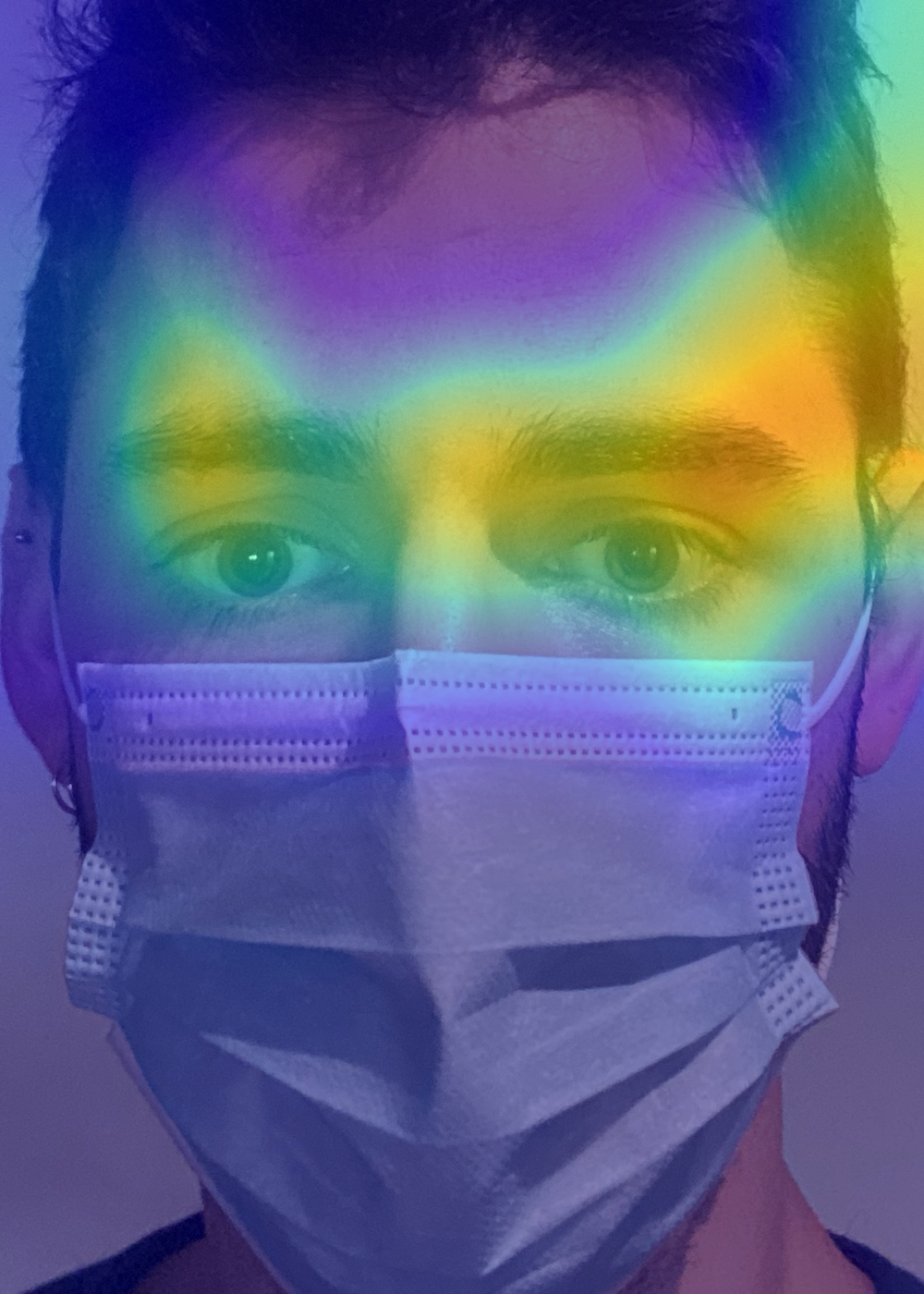}
       \caption{}
  \end{subfigure}
  \begin{subfigure}[b]{0.13\linewidth}
       \includegraphics[width = \linewidth,height=2.4cm]{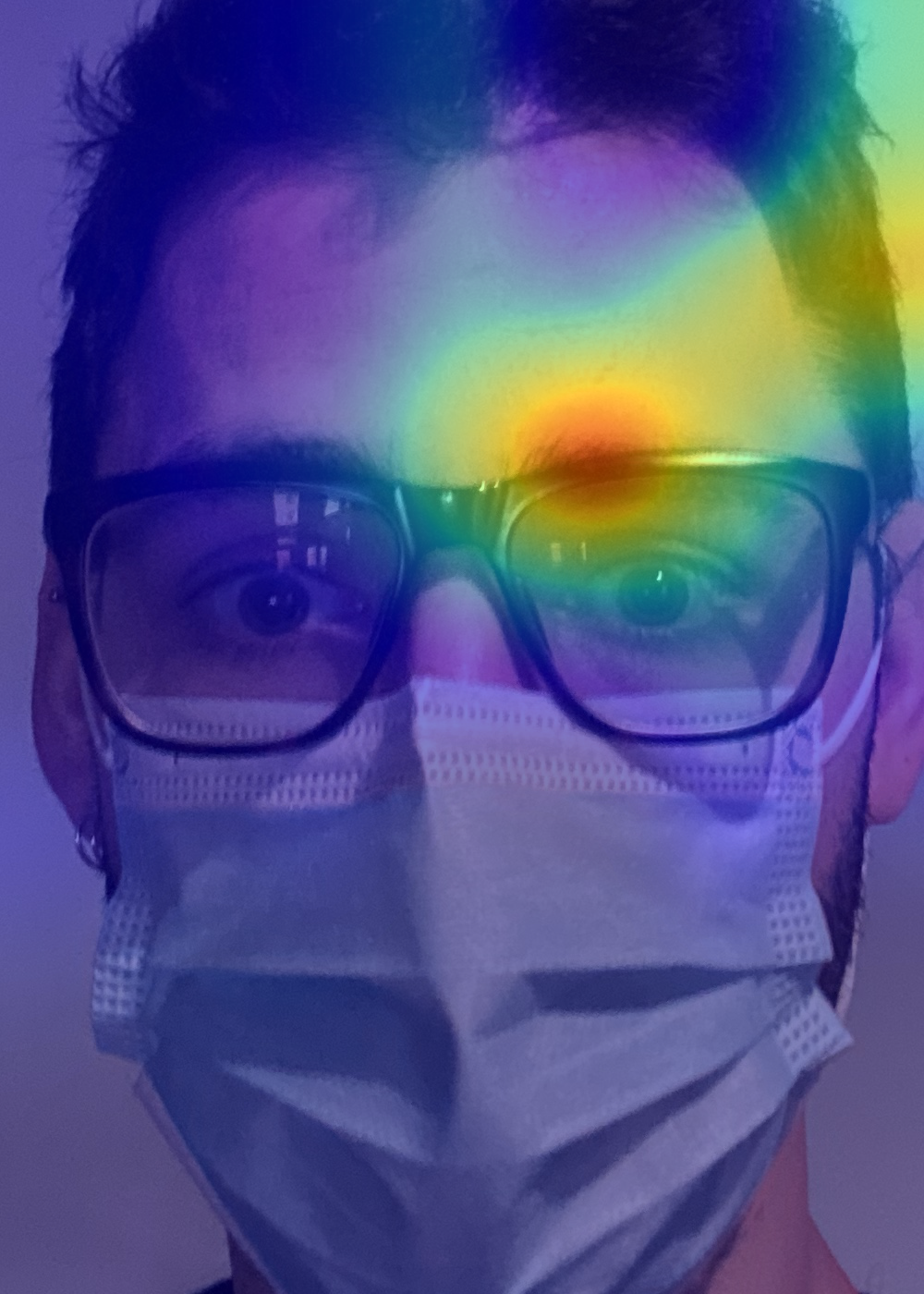}
       \caption{}
  \end{subfigure}
  
  \begin{subfigure}[b]{0.13\linewidth}
       \includegraphics[width = \linewidth,height=2.4cm]{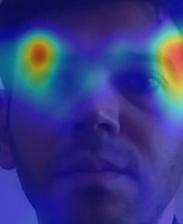}
       \caption{}
  \end{subfigure}
  \begin{subfigure}[b]{0.13\linewidth}
       \includegraphics[width = \linewidth,height=2.4cm]{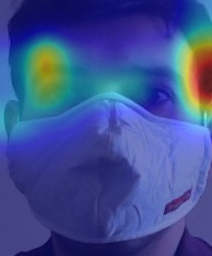}
       \caption{}
  \end{subfigure}
  \begin{subfigure}[b]{0.13\linewidth}
       \includegraphics[width = \linewidth,height=2.4cm]{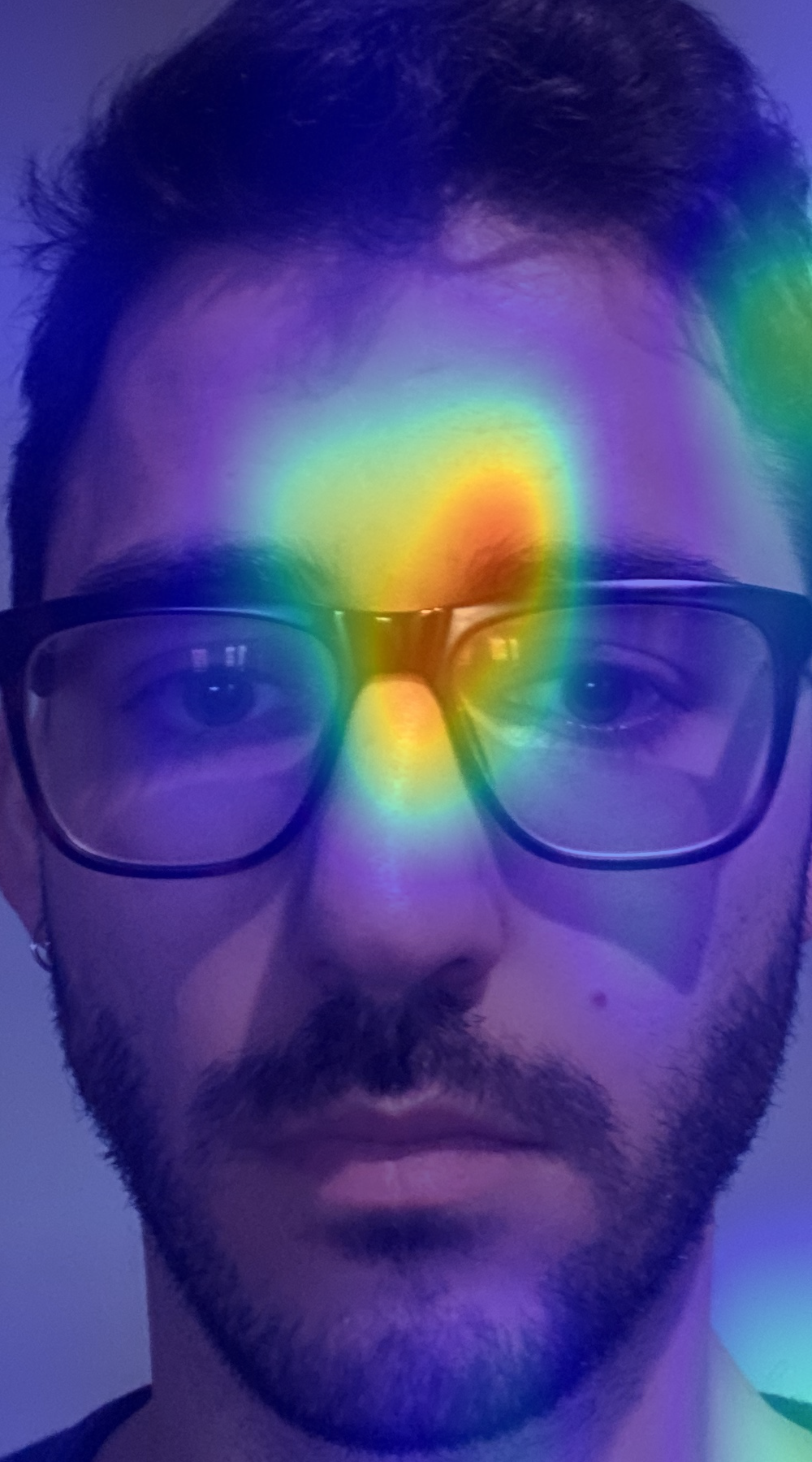}
       \caption{}
  \end{subfigure}
  \begin{subfigure}[b]{0.13\linewidth}
       \includegraphics[width = \linewidth,height=2.4cm]{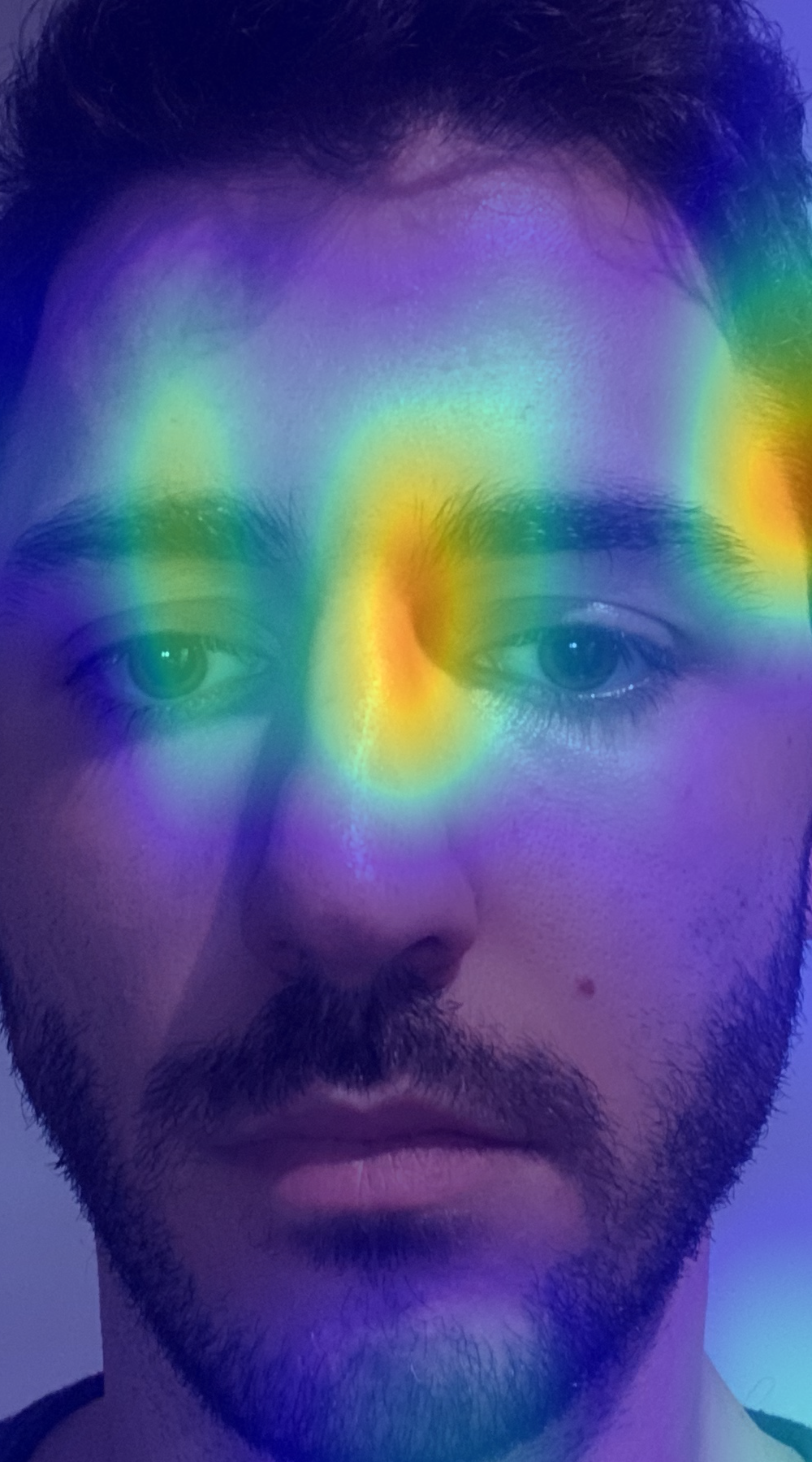}
       \caption{}
  \end{subfigure}
  \begin{subfigure}[b]{0.13\linewidth}
       \includegraphics[width = \linewidth,height=2.4cm]{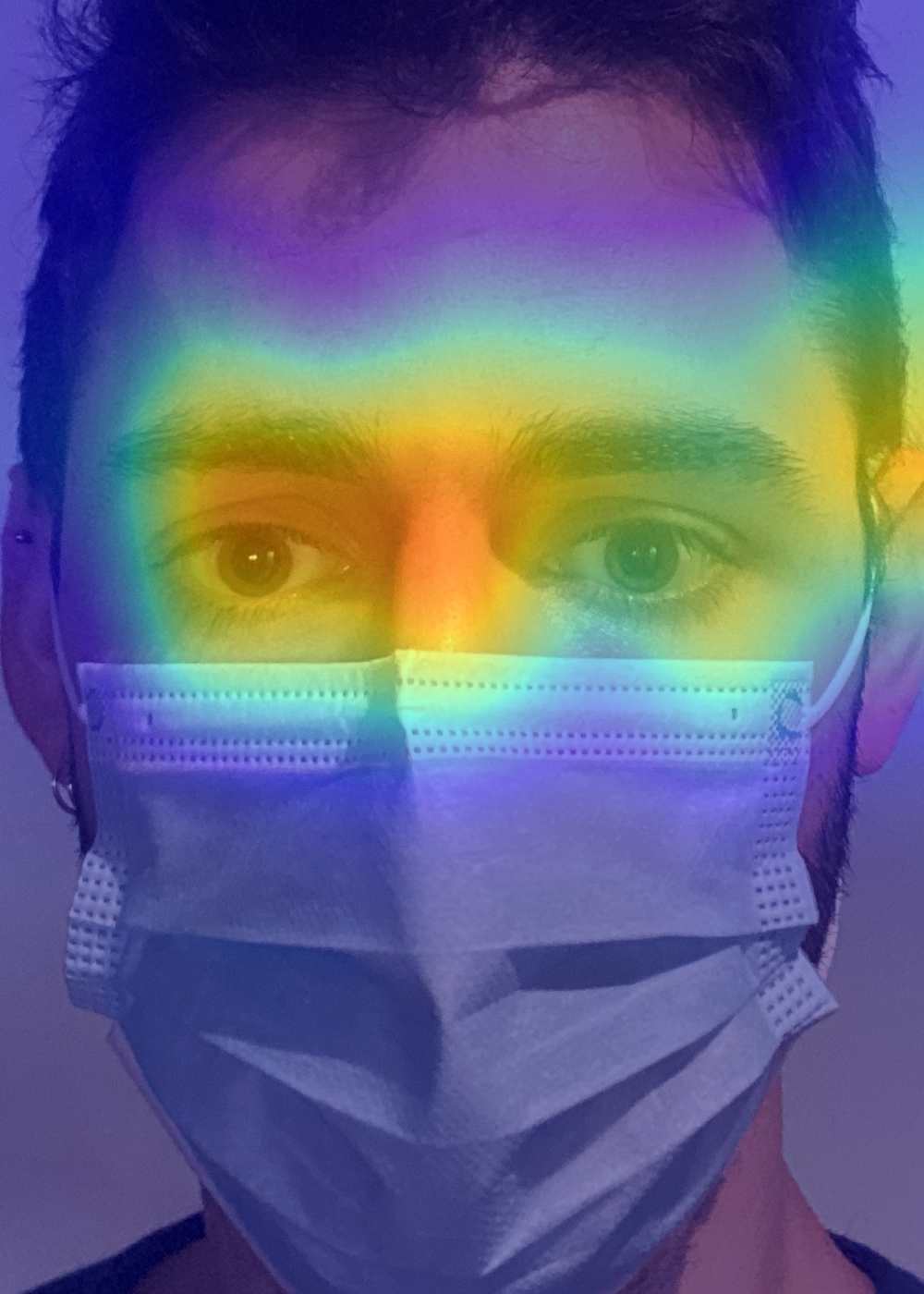}
       \caption{}
  \end{subfigure}
  \begin{subfigure}[b]{0.13\linewidth}
       \includegraphics[width = \linewidth,height=2.4cm]{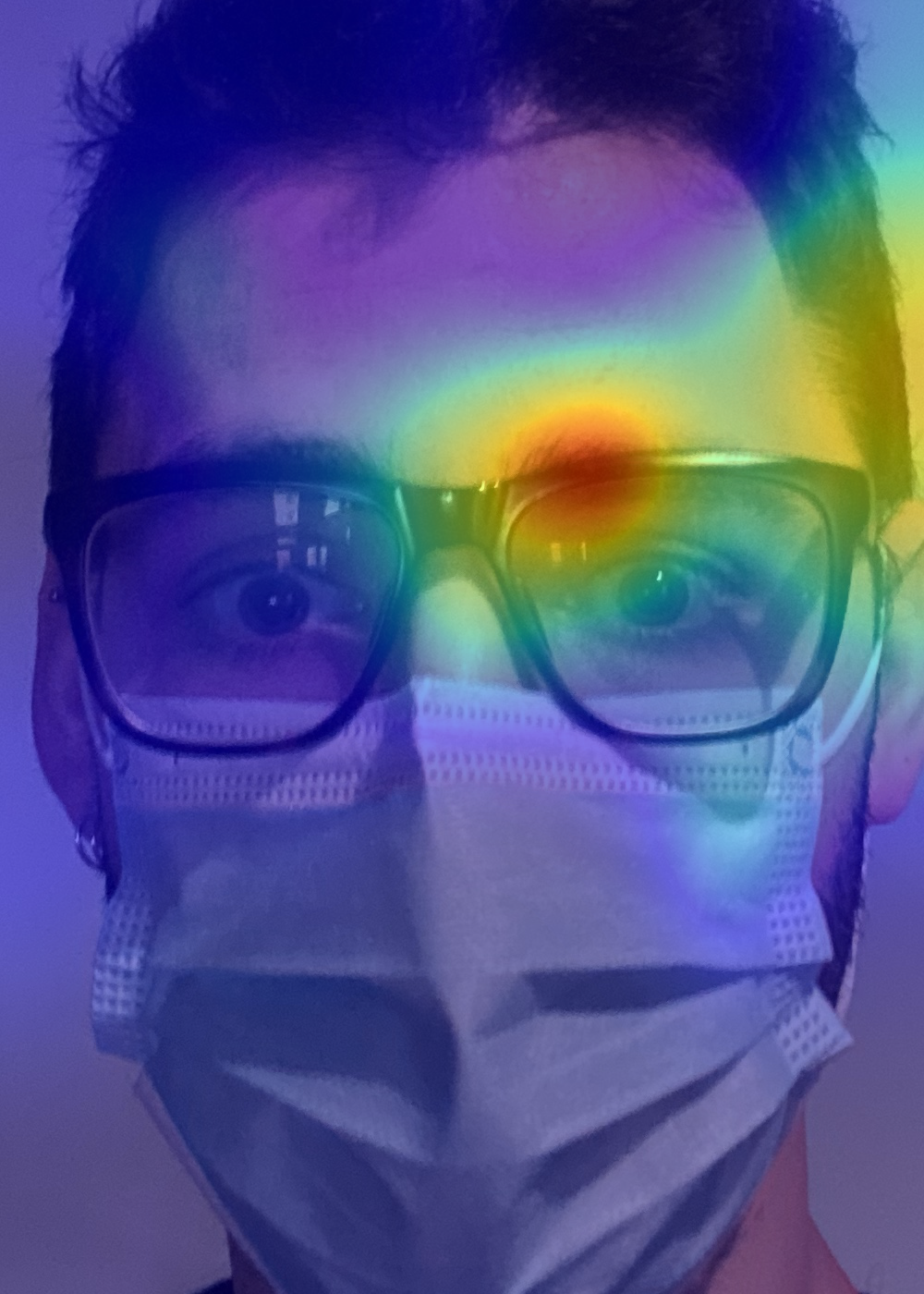}
       \caption{}
  \end{subfigure}
  \caption{{Output of the Smooth Grad-CAM++ computed for each of the 512 features of the feature vector and normalized. Subfigures (a)-(f) computed from the cross-entropy model; (g)-(l) computed from the triplet loss model; and (m)-(r) computed from the mean squared error and triplet loss model.}}
    \label{fig:grad_cam}

\end{figure}

Besides the quantitative evaluation of the proposed method, a more qualitative approach was also studied. In order to infer the importance of each pixel for the overall embedding produced we used the Smooth Grad-CAM++method. The output of this gradient-based method was computed for each of the 512 features. Afterwards, all the map outputs were summed and divided by 512, thus generating a final map with the average importance of the pixels to the overall embedding. Figure~\ref{fig:grad_cam} displays the outputs for three of the studied methods, from the top to the bottom we have the CE, the TL and the TL+MSE methods. While the first is already capable of ignoring the masks, it still constructs the embedding of the unmasked images based on the chin area. Between the other two models, the main difference seems to be that the model with the MSE uses a wider area of pixels to construct the embedding, thus, capturing more information.

\section{Conclusion and Future Work}
{In this work we addressed the challenge of masked face recognition motivated by the recent Covid-19 pandemic causing that wearing masks is now essential to prevent the spread of contagious diseases and has been currently forced in public places in many countries. However, recent research has shown that the performance, and thus the trust in contactless identity verification through face recognition, can be impacted by the presence of a mask. 
The scenario addressed is the evaluation of the verification performance in face recognition systems when verifying masked vs not-masked faces compared to verifying not-masked faces to each other. It was already noted in the literature, that the effect of masks was stronger on genuine pairs decisions in comparison to imposter pairs decisions. In this work, we proposed a methodology that targeted that observation and aimed at improving the performance of MFR systems in the comparison of unmasked versus masked faces. The results obtained by our proposed method showed consistent improvements in a detailed step-wise ablation study. The ablation studies performed showed that our proposed triplet loss modification improved the performance of the models in the addressed scenario. }


\section*{Acknowledgements}
{
This work was financed by National Funds through the Portuguese funding agency, FCT - Fundação para a Ciência e a Tecnologia within project UIDB/50014/2020, and within the PhD grants ``2021.06872.BD'' and ``SFRH/BD/137720/2018''. This research work has been also funded by the German Federal Ministry of Education and Research and the Hessen State Ministry for Higher Education, Research and the Arts within their joint support of the National Research Center for Applied Cybersecurity ATHENE.
\\}

\printbibliography
\end{document}

%% file: math_commands.tex
\usepackage{amsmath,amsfonts,amssymb,amsthm,bm}

\def\eqref#1{equation~(\ref{#1})}

\def\1{\bm{1}}

\DeclareMathAlphabet{\mathsfit}{\encodingdefault}{\sfdefault}{m}{sl}
\SetMathAlphabet{\mathsfit}{bold}{\encodingdefault}{\sfdefault}{bx}{n}

